\documentclass[letterpaper, 10 pt, conference]{ieeeconf}  

\IEEEoverridecommandlockouts                              

\usepackage{amsmath} 
\usepackage{amssymb}  
\usepackage[linesnumbered,ruled]{algorithm2e}
\usepackage{gensymb}
\usepackage{graphicx}
\usepackage{epstopdf}
\usepackage{subcaption}
\usepackage[colorlinks]{hyperref}
\usepackage{multirow}
\usepackage[T1]{fontenc}
\usepackage[super]{nth}
\usepackage{fix-cm}
\usepackage{floatrow}
\usepackage[]{caption}
\usepackage{lipsum}  

\makeatletter
\let\NAT@parse\undefined
\makeatother

\usepackage[square,comma,numbers,sort&compress]{natbib} 

\title{\LARGE \bf
Optimizing Terrain Mapping and Landing Site Detection for Autonomous UAVs
}

\author{Pedro F. Proen\c{c}a, Jeff Delaune, Roland Brockers$^{*}$
\thanks{$^{*}$ Jet Propulsion Laboratory, California Institute of Technology, \quad Pasadena, CA, USA}%
}

\begin{document}

\maketitle
\thispagestyle{empty}
\pagestyle{empty}

\begin{abstract}
The next generation of Mars rotorcrafts requires on-board autonomous hazard avoidance landing.
To this end, this work proposes a system that performs continuous multi-resolution height map reconstruction and safe landing spot detection.
Structure-from-Motion measurements are aggregated in a pyramid structure using a novel Optimal Mixture of Gaussians formulation that provides a comprehensive uncertainty model. Our multiresolution pyramid is built more efficiently and accurately than past work by decoupling pyramid filling from the measurement updates of different resolutions.
\par
To detect the safest landing location, after an optimized hazard segmentation, we use a mean shift algorithm on multiple distance transform peaks to account for terrain roughness and uncertainty. The benefits of our contributions are evaluated on real and synthetic flight data.

\end{abstract}


\section{Introduction}
NASA's Mars Helicopter, Ingenuity, lacks the ability to detect safe landing sites on unknown terrain, therefore flights have to be planned with a large safety margin and the terrain must be analyzed a-priori based on HiRISE and the rover's imagery. Likewise on Earth, commercial drones still rely heavily on \textit{return-to-home} mode and do not feature autonomous emergency landing. \par
To assess the landing safety of overflown terrain using a monocular camera, it is desired to reconstruct in real-time a multi-resolution Digital Elevation Map (DEM) to cope with different altitudes, high terrain relief, sensor noise and camera pitch. However multi-resolution maps raise issues in terms of computational cost. \par

Inspired by a Laplacian pyramid, previously we proposed to aggregate 3D measurements as residuals in a multi-resolution pyramid map for landing site detection \cite{Schoppmann2021multires}. To avoid updating all pyramid layers for each measurement, the pixel footprint is used as a criteria to select which layers to update.
Although this minimizes the number of pyramid updates for far away measurements (relative to the pyramid's resolution) it can lead to incoherent maps between layers as we demonstrate in this work.

\begin{figure}[t]
	\centering
	\includegraphics[width=\textwidth]{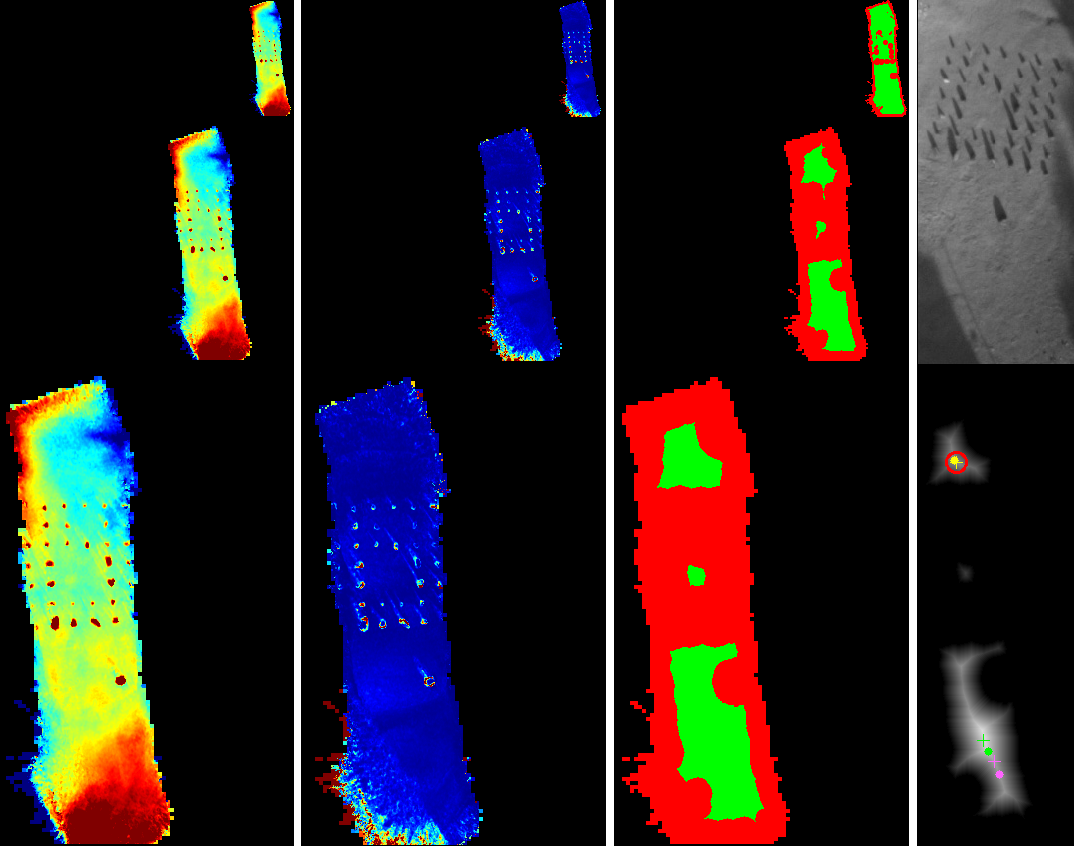}
	\caption{Multi-resolution maps with 3 layers (top-bottom) generated by our system along with landing spot detection during a 3 m altitude flight over a rock field (top-view shown on the top-right corner). Left to right:  DEM, DEM uncertainty and landing segmentation. Bottom-right shows the distance transform obtained from the binary landing segmentation and 3 landing spot candidates. These are initialized as distance transform peaks (dots) and then shifted (crosses) towards less locally rough and uncertain areas. Red circle marks the selected location. Notice how the uncertainty is higher on the rock discontinuities, outliers and shadows, where stereo noise is higher.}
	\label{fig:fig1}
\end{figure}

Here, we propose an alternative pyramid scheme that effectively decouples the mapping into 2 parts: a measurement update that selects the right layer and a pyramid filling process, called pyramid pooling, that enforces consistency.

Moreover, sensor fusion \cite{kweon1989terrain,forster2015continuous,fankhauser2014robot} has been dominated by Kalman updates (i.e. multiplication of Gaussians), including our past work \cite{Schoppmann2021multires}, where each pyramid cell is effectively a 1D Kalman filter. Although, optimal in the least squares sense, the Kalman filter uncertainty does not capture the discrepancy between measurements in a cell. Therefore, we propose fusing measurements using Optimal Mixture of Gaussians (OMG), by extending the work done in \cite{proencca2017probabilistic}. The advantage of OMG, demonstrated in Fig. \ref{fig:fig1} is that the uncertainty captures both the sensor error model and the observed variance between measurements within each cell.

To detect the safest landing spot, we improve \cite{Schoppmann2021multires} by introducing a light iterative solution that selects the location that minimizes the map roughness and uncertainty while maximizing the distance to closest hazards given by the map segmentation. Our results indicate this improves the landing selection under missegmented hazards.

\begin{figure*}[t]
\includegraphics[scale=0.65]{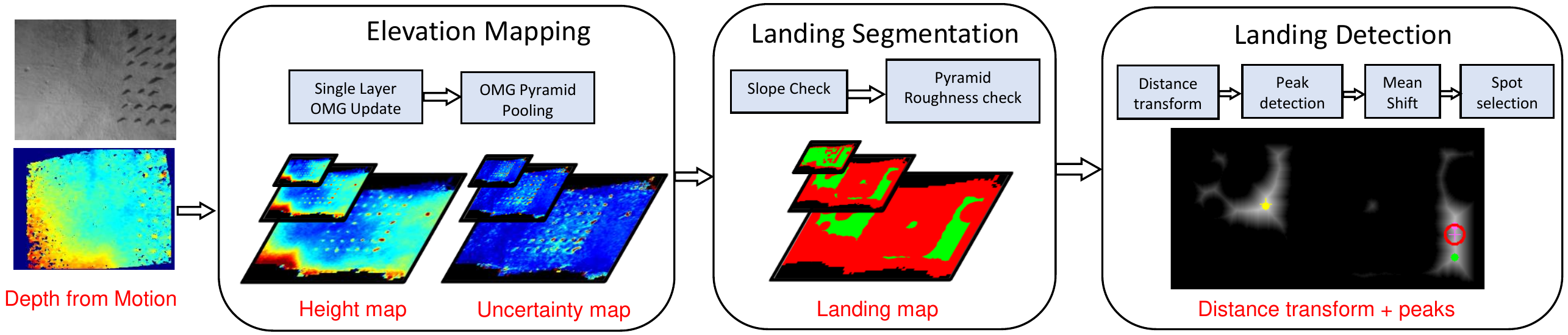}
\caption{System Pipeline. Point cloud measurements, given by visual-inertial odometry and stereo \cite{9438289}, are continuously aggregated in our pyramid structure. Then safe cells are segmented top-down by checking slope and roughness. Finally, we run a peak detection on the distance transform, refine these peaks and rank them based on roughness and uncertainty.}
\label{fig:pipeline}
\end{figure*}
In summary, we extend the work in \cite{Schoppmann2021multires} with the following contributions:
\begin{itemize}
\item A multi-resolution mapping scheme that decouples single-layer measurement updates from pyramid filling.
\item Fusing measurements using a new OMG cumulative form for an accurate uncertainty model.
\item Faster landing segmentation by using a rolling buffer during the roughness check.
\item An iterative landing detection method less sensitive to landing segmentation errors and parameter choices.
\end{itemize}

\section{Related Work}

The problem of terrain mapping for planetary exploration was initially addressed in \cite{kweon1989terrain}.
Since then, NASA has developed several hazard detection systems for spacecraft landing \cite{trawny2015flight,luna2017evaluation,johnson2002lidar,Johnson20205icra}. However \cite{trawny2015flight,luna2017evaluation,johnson2002lidar} use a flash LiDAR, which is not suitable for weight-restricted UAVs (e.g. Mars Helicopter). LiDAR was also used in \cite{Scherer2012heli} to assess the terrain for landing a Helicopter by considering the rotor and tail clearance.

Recently, the Mars 2020 mission used for the first time visual Terrain Relative Navigation (TRN) \cite{johnson2017lander} to land on a more hazardous yet more scientifically interesting location than past Mars landing locations. TRN estimates the spacecraft location with respect to a HiRISE map annotated \textit{a-priori} with large Hazards. The HiRISE resolution is however not enough for small UAV hazards. Closer to our work, \cite{Johnson20205icra} proposed an algorithm for landing hazard avoidance based on a moving monocular camera. In that work, a dense DEM is reconstructed by interpolating measurements from a single frame but measurements are not aggregated across multiple frames. The DEM is binarized based on plane fitting and a simple Distance Transform is then used to select the safest location closer to an \textit{a-priori} landing site.

Continuously fusing measurements with a single layer DEM was done in \cite{fankhauser2014robot,forster2015continuous,fankhauser2016universal} using Kalman updates. Although \cite{forster2015continuous} proposed a mechanism to switch between map resolutions for different altitudes, this is not flexible enough for 
complex 3D terrain and off-nadir viewpoints. In \cite{triebel2006multi} elevation maps are extended to represented multiple surface levels, common in urban environments. For such structured scenes, \cite{bosch2006autonomous,brockers2011autonomous} proposed using Homography to detect and land on large planes (e.g. rooftops). In \cite{hinzmann2018free,warren2015enabling} landing sites are detected based on both texture and geometric shape.
Gaussian Processes have been used in \cite{vasudevan2009gaussian} to interpolate sparse maps and in \cite{popovic2017multiresolution} for multiresolution mapping. In \cite{Schoppmann2021multires} we proposed a multi-resolution hazard mapping and segmentation approach. This work extends \cite{Schoppmann2021multires} in several ways as discussed in the next section.

\section{System Overview}
\label{sec:mapping}

Our system, illustrated in Fig. \ref{fig:pipeline}, starts by aggregating point clouds, generated per image, into a pyramid structure representing a multi-resolution DEM. These point clouds are obtained, as in \cite{Schoppmann2021multires}, by relying on the Structure-from-Motion system proposed in \cite{domnik2021dense}, which couples a range-visual-inertial odometry \cite{delaune2020b}, a local Bundle Adjustment and a standard dense stereo algorithm. Also, as in  \cite{Schoppmann2021multires}, to keep up with the UAV motion the map is shifted using a  rolling buffer  if necessary to fit the observed terrain. \par

\begin{figure}[tb]
	\centering
	\begin{tabular}{@{}c|c@{}}
	\includegraphics[scale=0.5]{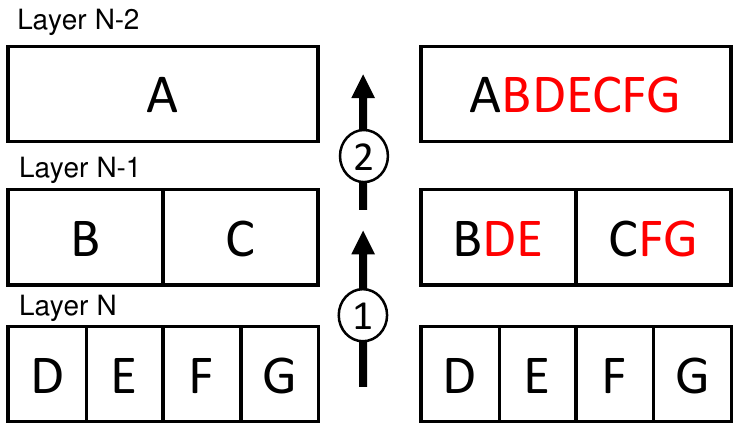} &
	\includegraphics[scale=0.5]{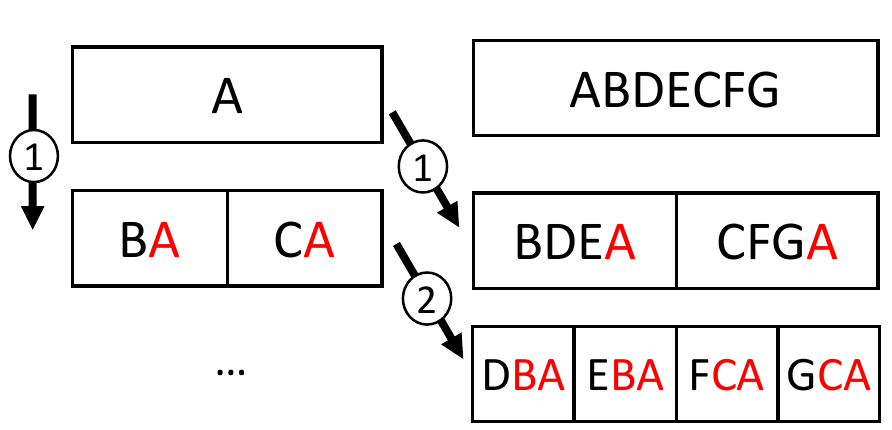} \\
	(a) Up-Pooling & (b) Down-Pooling
	\end{tabular}
	\caption{Pyramid pooling scheme illustrated for a 1D \textit{cell's pyramid}  example -- covering the terrain footprint of the top cell. Letters represent the cell's states and arrow numbers represent order of fusion. (a) \textit{Left} represents the input pyramid (after the single layer updates). (b) \textit{Left} is a copy of the input pyramid used for down-pooling. \textit{Right} is the output. If a target cell is empty, the source cell is simply copied.}
	\label{fig:pyramid_pooling}
\end{figure}

The DEM pyramid structure consists of N layers. The resolution of a layer is half of the resolution of the layer below, i.e., the terrain footprint covered by one cell at the top layer is the footprint of 4 cells in the layer below. Hereafter, we will simply refer to all the cells covering that top cell's footprint as \textit{cell's pyramid}. This is illustrated in Fig \ref{fig:pyramid_pooling}. In \cite{Schoppmann2021multires}, each cell is a 1D Kalman Filter state containing 2 values: fused height (or height residual) and variance. In our case each cell is a 3-value OMG state, described in Section \ref{sec:OMG}.

Once we get a new point cloud, each point measurement updates only one cell from one selected layer. We select the layer where the measurement's footprint fits the best, specifically, the lowest layer where the cells' footprint is larger than the measurement's footprint given by $z/f$ where $z$ is the measurement's depth an $f$ is the camera focal length. This criteria was used in \cite{Schoppmann2021multires} but to update all layers where the cell's footprint is larger than the measurement's footprint. Then, we select the cell by quantizing the ground plane coordinates of the measurement given the resolution of the selected layer.

After updating all new measurements, we can recover the complete \textit{cell's pyramids} in a separate module called Pyramid Pooling, explained in Section \ref{sec:pooling}, where the cell states are fused across the respective \textit{cell's pyramid}. This produces the same result as if we instead updated directly all layers for every single measurement but with significantly less updates and cache misses.

  We use the same coarse-to-fine cascade segmentation approach as in \cite{Schoppmann2021multires}: First, we check the terrain roughness and slope at the top layer and then check only the roughness consecutively at the lower layers. The slope is obtained by fitting planes to the top cells and their neighbourhood within a circular area, given a user defined landing radius based on the UAV size. The roughness involves searching the minimum and maximum height within another user-defined circular neighbourhood. In Section \ref{sec:seg}, we propose an optimization to this roughness check by reducing the min and max search. Cells are labeled as safe if the max roughness within the landing radius is smaller than a roughness threshold.

For landing detection, detailed in Section \ref{sec:lsd}, we use the binary landing map to compute the distance transform, which gives the distance to the closest landing hazard, where we then detect the maximum N peaks. These peaks are then subject to a mean shift based on the roughness map, which was used during the segmentation. Finally, we can select the best landing location based on a variance criteria around the landing area.

\section{Infinite Optimal Mixture of Gaussians}
\label{sec:OMG}
Given M measurements $\{x_1,...,x_M\}$ and their uncertainties $\{\sigma^2_{x_1},...,\sigma^2_{x_M}\}$
The probability density function for an Optimal Mixture of Gaussians is defined as
\begin{equation}
f(x) = \frac{1}{S} \sum_{n=1}^{M} \sigma_{x_i}^{-2} N(x_i,\sigma_{x_i}^{2})
\textrm{,} \quad S = \sum_{n=1}^{M} \sigma_{x_i}^{-2}
\end{equation}
In our case, $x_i$ and $\sigma_{x_i}^2$ are a height value and uncertainty. For simplicity, in this work, $\sigma_{x_i}^2$ models only the stereo depth quadratic noise due to disparity quantization. As derived in \cite{proencca2017probabilistic}, the resulting OMG mean and variance is:

\begin{equation}
\mu = \frac{1}{S}\sum_{n=1}^{M} \frac{x_i}{\sigma_{x_i}^{2}}
\textrm{,} \quad
\sigma^2 = \frac{1}{S}\sum_{n=1}^{M} \frac{\sigma_{x_i}^{2}+x_i^2}{\sigma_{x_i}^{2}}- \mu^2
\label{eq:OMG_ori}
\end{equation}

The mean is identical to a Kalman Filter (See Appendix \ref{sec:appendix}) giving us a Maximum Likelihood Estimation. However the OMG variance allows us to capture the variance between measurements within a cell besides the measurement prior uncertainties.

The formulation (\ref{eq:OMG_ori}), used in \cite{proencca2017probabilistic}, is not scalable as this would require storing all measurements, but we can rearrange this to a cumulative form,
where at every time step $t$, a new measurement $x_t$ with $\sigma_{x_t}^{2}$ updates the prior OMG state $\{\mu_{t-1},\sigma^2_{t-1},S_{t-1}\}$ using: 
\begin{equation}
S_t = S_{t-1} + \sigma_{x_t}^{-2}
\label{eq:S_cum}
\end{equation}
\begin{equation}
\mu_t = \frac{1}{S_t} \left(S_{t-1}\mu_{t-1} + \frac{x_t}{\sigma^2_{x_t}}\right)
\label{eq:mean_cum}
\end{equation}
\begin{equation}
\sigma^2_t = \frac{1}{S_t} \left(S_{t-1}(\sigma^2_{t-1}+\mu^2_{t-1}) + \frac{x^2_t}{\sigma^2_{x_t}}+1\right)-\mu^2_t
\label{eq:var_cum}
\end{equation}
Therefore, each cell state needs to store $\{\mu_{t},\sigma^2_{t},S_{t}\}$. It's worth noting that although we did not suffer from numerical errors in our experiments, the squared terms involved in the last expression: $x^2_t$ and $\mu^2_t$ may lead to overflow for very large height values $x_t$. This can be prevented for example by first dividing all terms in (\ref{eq:var_cum}) by $x_t$ and then multiplying the result of  (\ref{eq:var_cum}) by $x_t$.

Both Kalman and OMG updates will converge after many measurements, which can be a problem in certain descent flights where high altitude measurements dominate the \textit{cell's pyramid}. Thus, we introduce an optional time inflation operation, where in each frame we can effectively multiply the past measurements' uncertainty by a factor $k$ just by replacing the OMG state $\sigma^2$ and $S$ by:
\begin{equation}
\sigma^2_{I} = \sigma^2 + (N_t+1)(k-1) 
\quad \textrm{and} \quad
S_I = S/k
\end{equation}
where $k$ is the inflation factor, $N_t$ is the number of measurements at time $t$.

\section{Pyramid Pooling}
\label{sec:pooling}
To obtain our final pyramid map, for each measurement we could simply use (\ref{eq:S_cum}), (\ref{eq:mean_cum}) and (\ref{eq:var_cum}) to directly update the cells of all layers that cover the measurement's footprint, but here we propose an indirect approach that achieves the same results with less updates in two stages. First, as discussed in Section \ref{sec:mapping}, for each measurement, we update only a cell from a selected layer using (\ref{eq:S_cum}), (\ref{eq:mean_cum}) and (\ref{eq:var_cum}). 

Then, as illustrated in Fig. \ref{fig:pyramid_pooling}.a, we fuse the states of the \textit{cell's pyramid} layer by layer, first from bottom to top by updating the upper layer for each filled cell in the lower layer and then, as shown in Fig. \ref{fig:pyramid_pooling}.b from top to bottom by updating the lower layer -- while using a copy of the original pyramid. We minimize cache misses by accessing sequentially the memory of all 2D arrays, therefore states are fused one by one into the target state. We can do so by relying on the associative property of the OMG fusion (\ref{eq:OMG_ori}), i.e., fusing a group of measurements is equivalent to split these measurements into two groups, fuse the groups individually and then fuse the two resulting states. Without loss of generality, let $\{\mu_{A},\sigma^2_{A},S_{A}\}$ and $\{\mu_{B},\sigma^2_{B},S_{B}\}$ be two cells, then we can fuse them using:
\begin{equation}
S_{A+B} = S_A + S_B
\label{eq:S_cum_meta}
\end{equation}
\begin{equation}
\label{eq:mean_cum_meta}
\mu_{A+B} = \frac{1}{S_{A+B}} \left(S_{A}\mu_{A} + S_{B}\mu_{B}\right)
\end{equation}
\begin{equation}
\label{eq:var_cum_meta}
\sigma^2_{A+B} = \frac{1}{S_{A+B}} \left(S_{A}(\sigma^2_A+\mu^2_{A}) + S_{B}(\sigma^2_B+\mu^2_{B}) \right)-\mu_{A+B}^2
\end{equation}

As demonstrated in Section \ref{sec:experiments}, both up-pooling and down-pooling leads to fewer updates than a naive multi-layer direct update. 

\section{Optimized Hazard Segmentation}
\label{sec:seg}
The computation cost of the segmentation process in \cite{Schoppmann2021multires} is dominated by the roughness check at the lower layers. Formally, roughness is computed for each cell as: $\textrm{max}_{i\in \Omega}h_i - \textrm{min}_{i\in \Omega}h_i$ where $h_i$ is the height at $i$th cell and $\Omega$ is the set of all cells within a distance to the query cell. \par
\begin{figure}[t]
	\centering
    \includegraphics[scale=0.65]{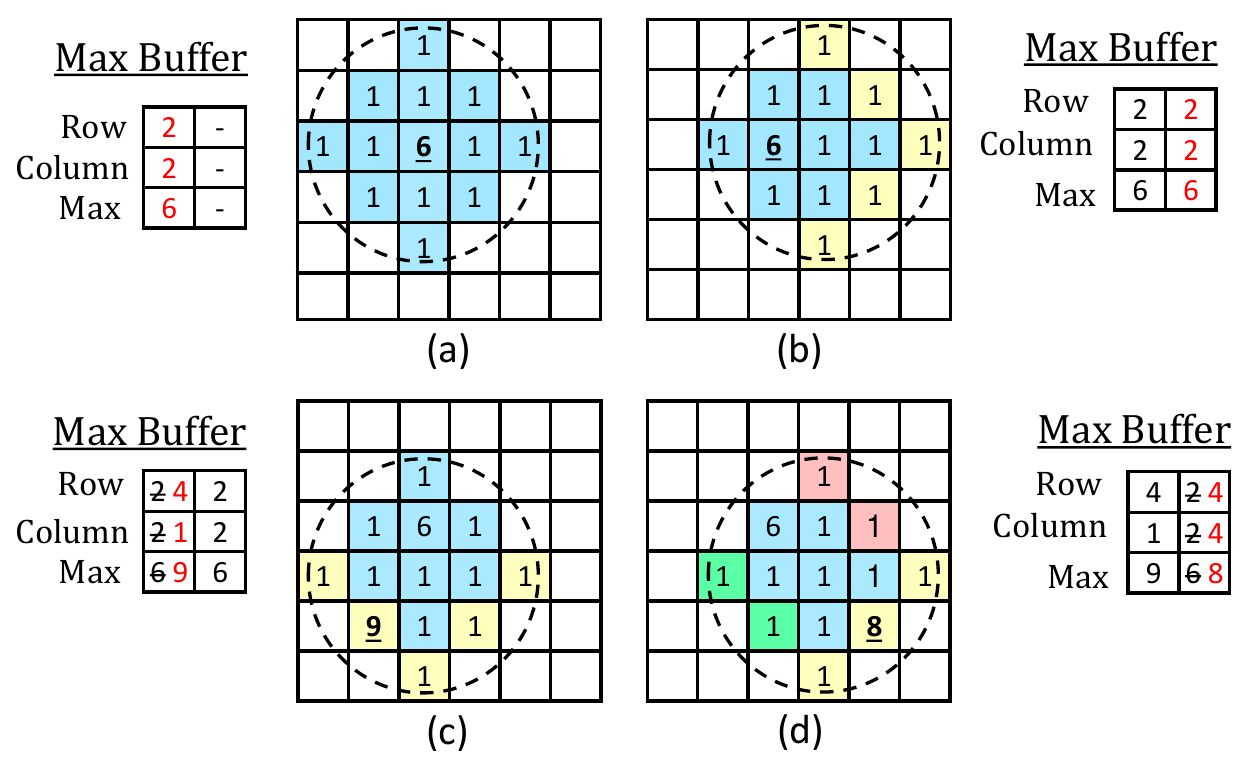}
	\caption{Example of Max search using a rolling buffer for roughness computation. Roughness is computed alphabetically from (a) to (d) for the cells in the circle center. (a) Initial cell, max needs to be searched in the blue region. (b) The coordinates stored in the last buffer's entry are within the search region so we only need to search the yellow region. (c) Similar to (b). (d) We only need to search the green and yellow region.}
	\label{fig:roughness_check}
\end{figure}

As shown in Fig. \ref{fig:roughness_check}, in certain cases we do not need to search the min and max for the entire circular region $\Omega$, since most of that region overlaps with past searched regions from cell neighbours. We exploit this fact by storing the min and max values and their 2D locations in a 1D rolling buffer (row-major). This leads to 4 search subregions, highlighted in Fig. \ref{fig:roughness_check}.d, that depend on the status of top and left cell, i.e., if the stored max and min locations are still within the new $\Omega$? If yes, then we can skip searching 2 or 3 subregions.

\section{Landing Site Detection}
\label{sec:lsd}
To select the safest landing location \cite{Schoppmann2021multires} and \cite{Johnson20205icra}
aim to maximize the distance from the selected location to its closest segmented hazard by computing the distance transform \cite{borgefors1986distance} from the binary segmentation map and then simply select the cell with the largest distance.

However this simple method assumes the segmentation is perfect and is sensitive to the roughness threshold. In reality, missegmented rocks start to show as the map resolution decreases, as reported in \cite{9438289}. Therefore, we propose a landing site detection method that weights the roughness and uncertainty map -- ignored in \cite{Schoppmann2021multires}.

The key idea is to shift that selected location, given by \cite{Schoppmann2021multires}, to a nearby location that minimizes the roughness and uncertainty within the landing area. This is done by running a Mean Shift algorithm that computes for a few iterations the shifted mean (location) using:
\begin{equation}
\label{eq::ms}
u_{t+1} = \frac{1}{\sum_{i\in \Omega} K(\phi(p_i))} \sum_{i\in \Omega} K(\phi(p_i))p_i
\end{equation}
where $\Omega$ is the sampled region around the current location $u_t$, $\phi(p_i)$ is the projection of cell's coordinates $p_i$ into feature space (e.g. roughness) and $K$ is a Gaussian Kernel:
\begin{equation}
K(x) = e^{-x^T\Lambda x} 
\label{eq:kernel}
\end{equation}
where $\Lambda$ is a diagonal matrix containing feature weights. In our case, we used $\phi(p_i)=[R_{i}, 1-D_{i}, \sigma_{i}]$ where $R_i$, $D_i$ are respectively the roughness and the normalized distance transform (i.e. $D\in[0,1]$)  at the $i$th cell.
Embedding the distance in the feature space acts as a regularization to prevent moving too far from the distance transform ridge, as shown in Fig \ref{fig:fig1}. Once we calculate (\ref{eq::ms}) we sample again $\Omega$ and repeat the process for a few iterations.

Because the mean shift is prone to local minima and we might have separate safe landing regions with similar distance transforms. We actually perform first a multiple peak detection on the distance transform and then do the Mean Shift individually for each peak. The peak detection uses Non-Maximum Suppression and selects up to $N$ peaks that are at least larger the largest peak scaled by a factor. Then we fit an OMG to the landing area using (\ref{eq:S_cum_meta}),(\ref{eq:mean_cum_meta}) and (\ref{eq:var_cum_meta}) around each peak using the uncertainty map and finally select the landing spot that has the smallest OMG variance.

\begin{figure*}
\centering
\begin{tabular}{@{}c@{ }c@{}}
\includegraphics[scale=0.31]{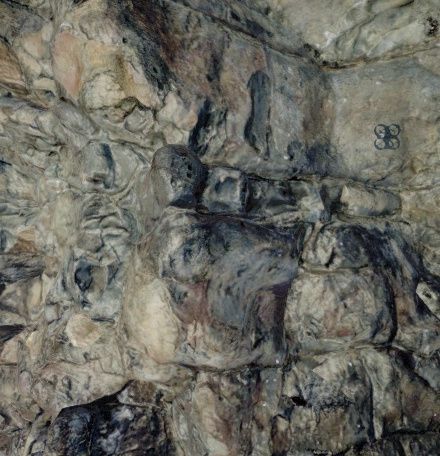} &
\includegraphics[scale=0.13]{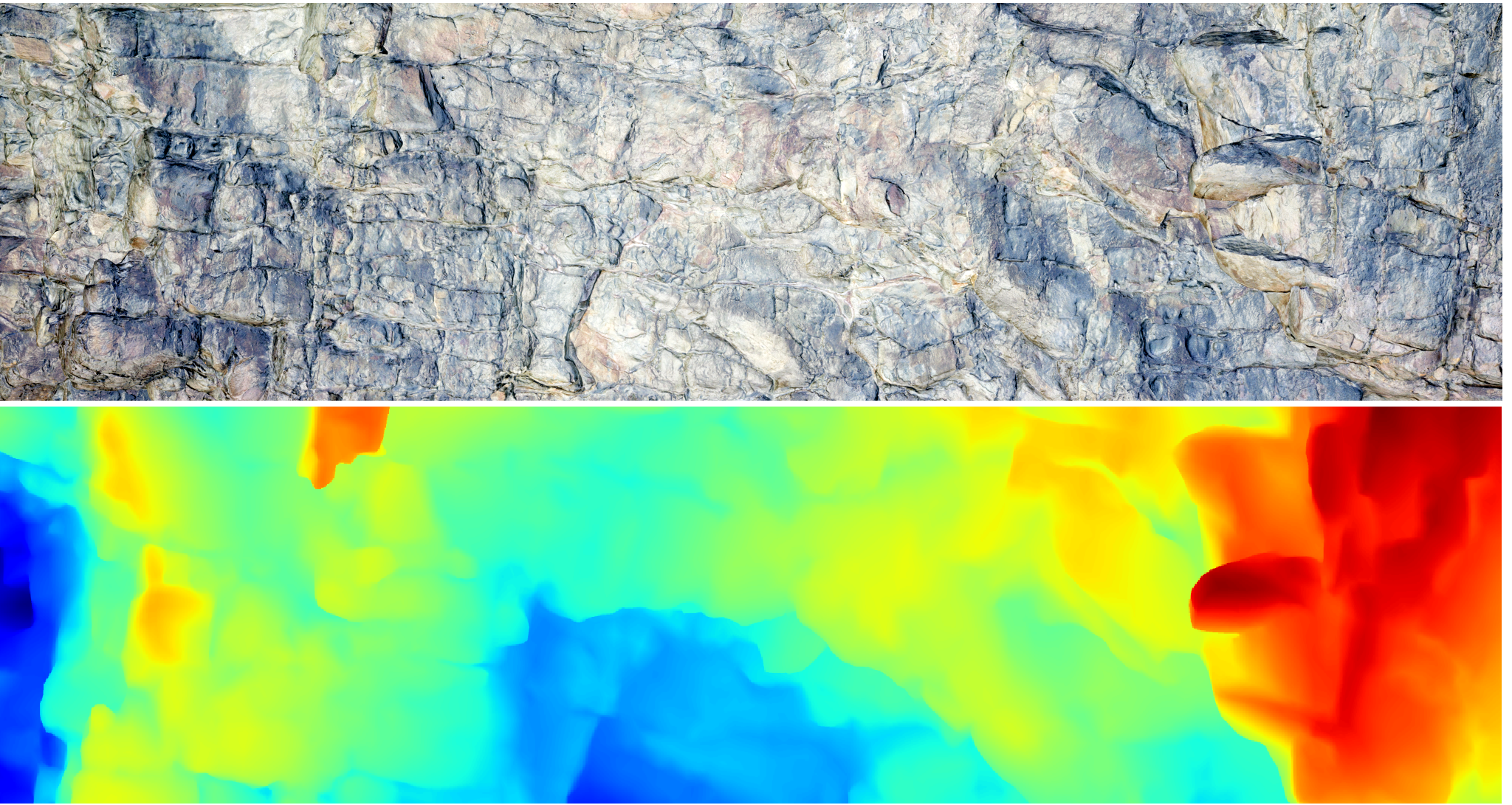} \\
(a) & (b) \\
\end{tabular}
\vspace{-10pt}
\caption{Synthetic terrain used in our experiments. A flight was executed flying 60 m from left to right in (b) and descending 10 m. (a) A frame example from the flight data. (b) RGB and Height orthomaps.}
\label{fig:synth_terrain}
\end{figure*}

\begin{figure}[tb]
	\centering
	\begin{tabular}{@{}c@{\hspace{3pt}}c@{}}
	\includegraphics[angle=90,scale=0.15]{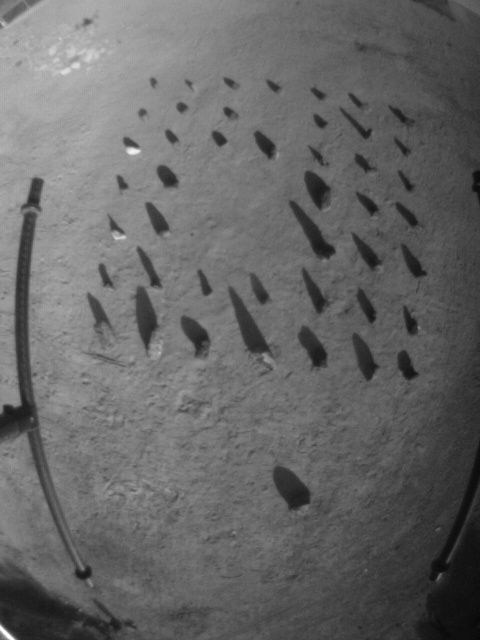} &
	\includegraphics[angle=90,scale=0.15]{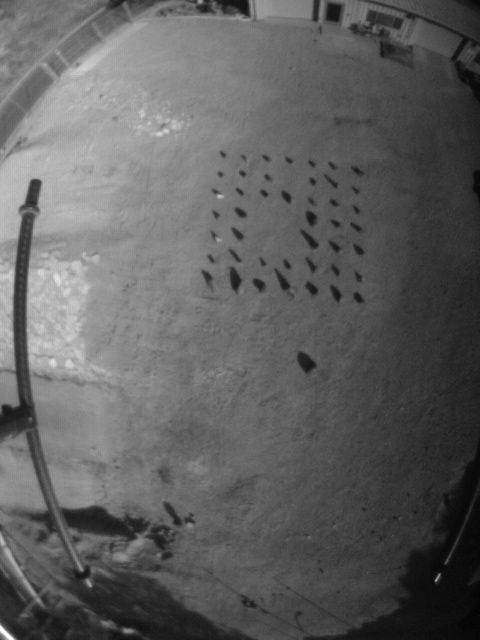}
	\vspace{-10pt}
	\end{tabular}
	\caption{Frames from the flights over the Mars Yard. Altitude above ground: 5 m (left) and 10 m (right).}
	\label{fig:my_flights}
\end{figure}

\section{Experiments}
\label{sec:experiments}

We evaluated our system modules on both real and synthetic datasets. For synthetic data, we used Unreal Engine and AirSim \cite{shah2018airsim} to render the scene shown in Fig. \ref{fig:synth_terrain}, during a descending trajectory. A global DEM is also generated to evaluate the reconstruction errors per frame. For real data, we used the flight data, shown in Fig. \ref{fig:my_flights}, that was collected during a flight campaign \cite{9438289} at the JPL Mars Yard. Multiple flight were performed at different altitudes above ground.
Based on previous fine-tuning segmentation results \cite{9438289}, except where explicitly mentioned, we set the following default parameters: For mapping, map resolution of lowest layer: 5 cm/cell, map size: 24$\times$24 m, number of layers: 3 and the uncertainty of first measurement in the cell is inflated by 25. For segmentation, maximum roughness threshold: 0.1 m, landing radius: 0.5 m, roughness search region radius: 0.5 m. For landing detection, maximum number of detected peaks: 5, number of mean shifts: 5, and, in (\ref{eq:kernel}), we used roughness weight: 100, distance weight: 10, uncertainty weight: 100.

\subsection{Mapping results}

\begin{figure}
	\centering
	\begin{tabular}{@{}c@{\hspace{3pt}}c@{}}
	\includegraphics[scale=0.42]{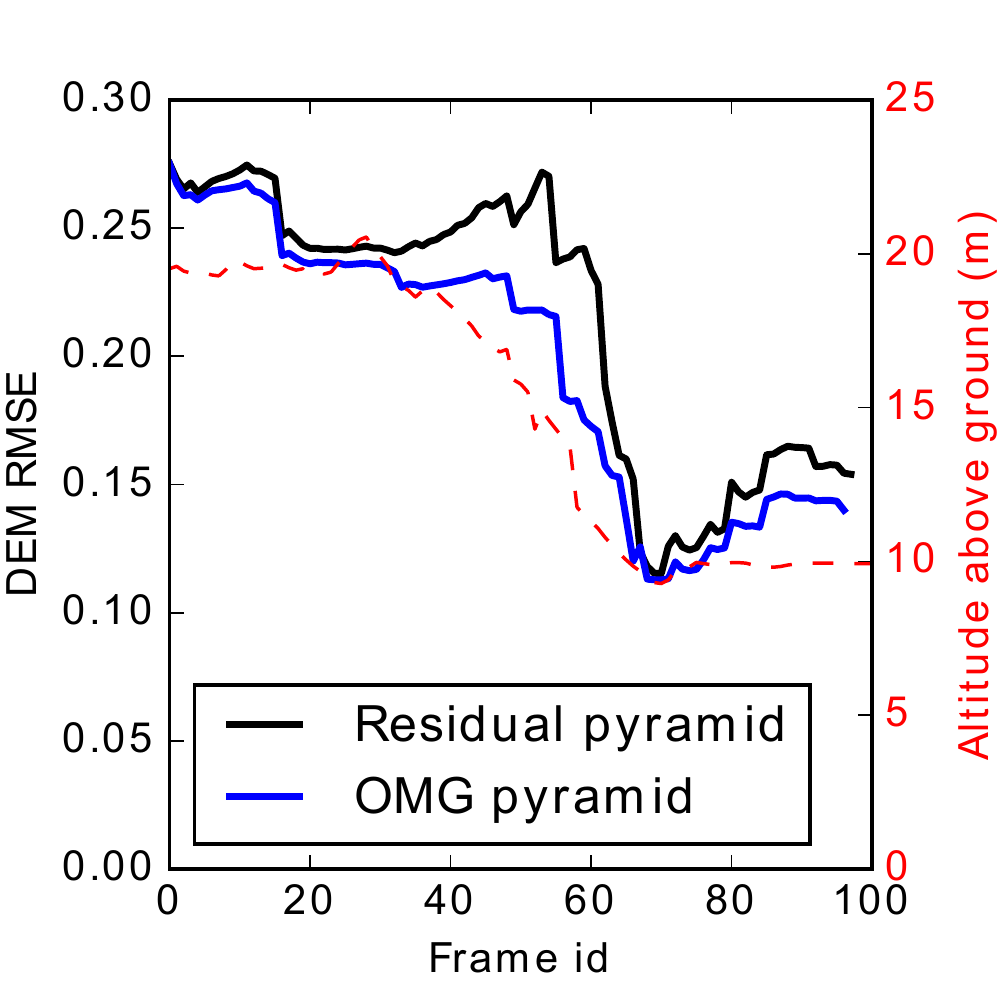} &
	\includegraphics[trim={0 15cm 14.5cm 0},clip,scale=0.42]{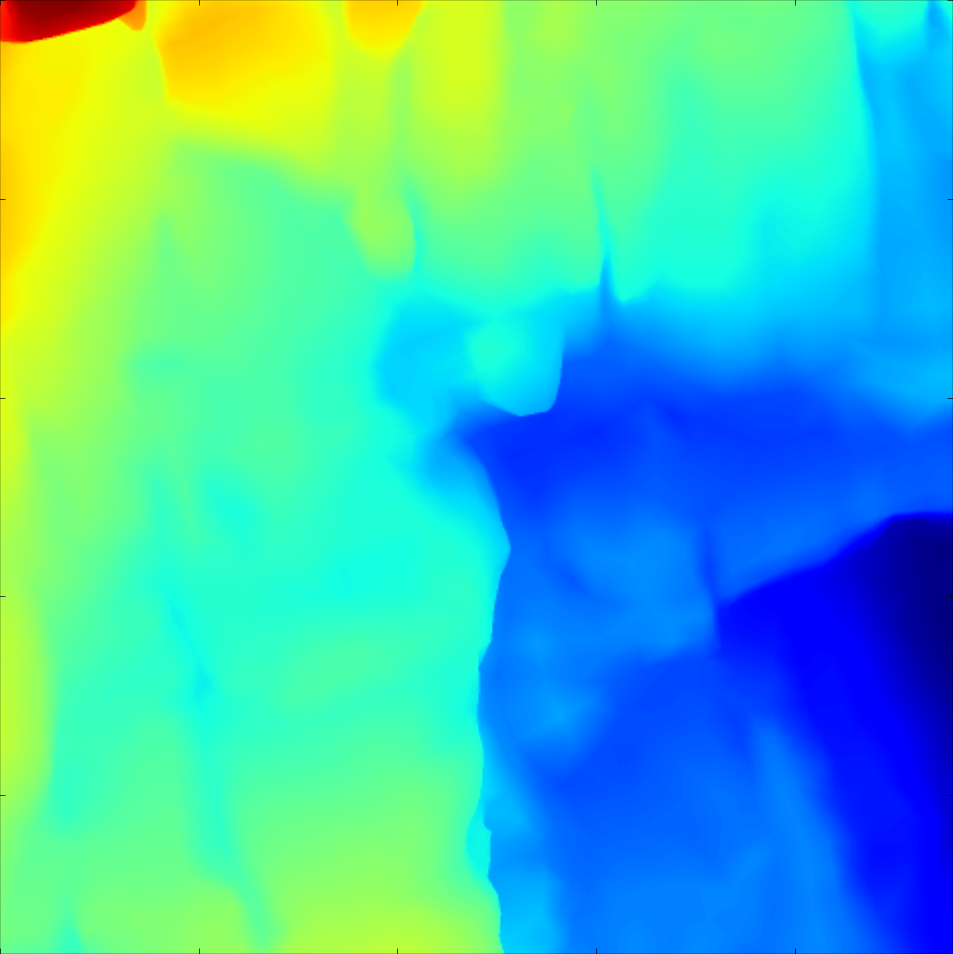} \\
	 \scriptsize{(a) Map RMSE per frame} & \scriptsize{(b) Ground truth DEM}\\
	\vspace{-5pt}
	\\
	\includegraphics[trim={0 15cm 14.5cm 0},clip,scale=0.42]{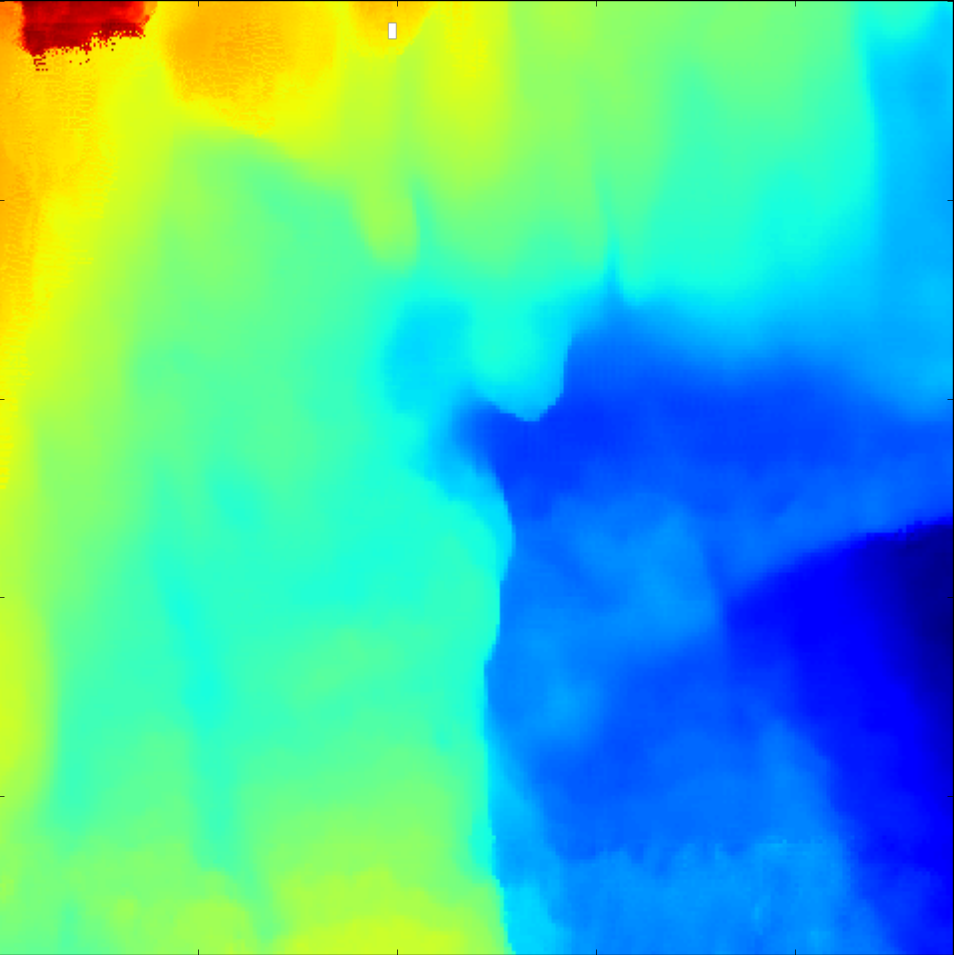} &
	\includegraphics[trim={0 15cm 14.5cm 0},clip,scale=0.42]{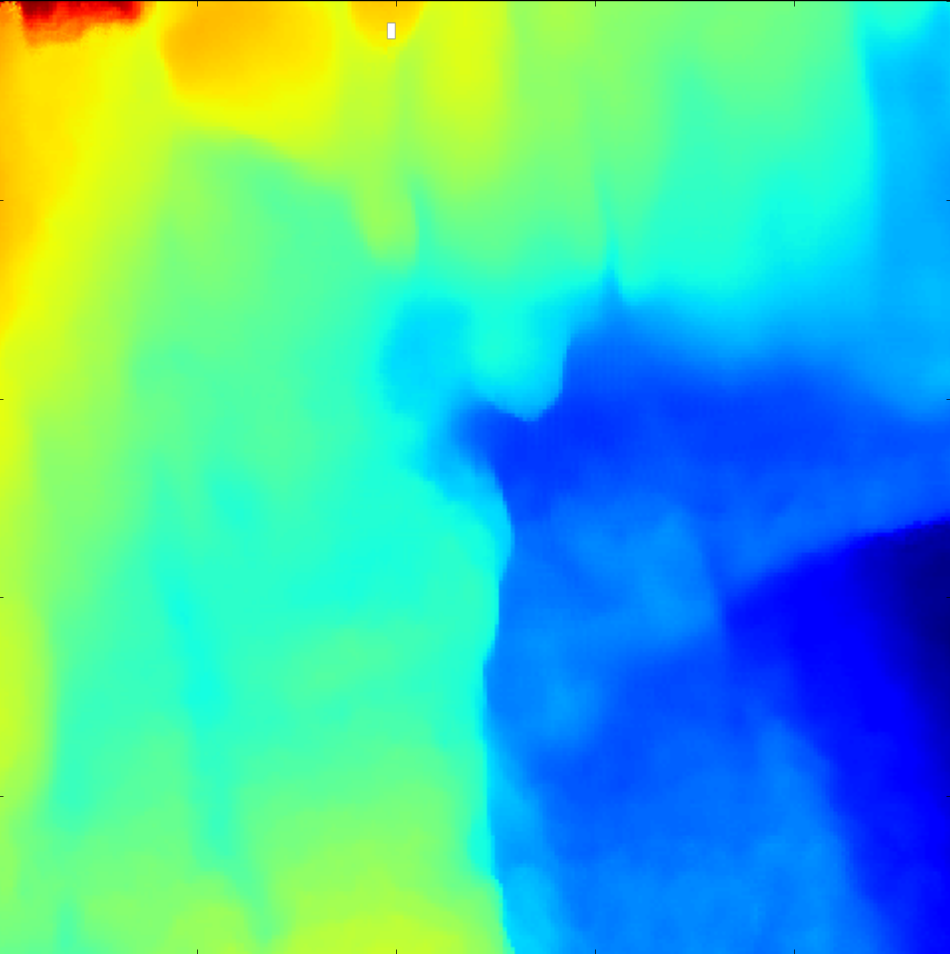} \\
	\scriptsize{(c) DEM using Residual Pyramid} & \scriptsize{(d) DEM using OMG Pyramid }\\
	\vspace{-15pt}
	\end{tabular}
	\caption{Mapping accuracy on our synthetic dataset (Fig. \ref{fig:synth_terrain}). (a) RMSE of the entire local DEM per sequence frame of the residual pyramid \cite{Schoppmann2021multires} vs our approach. The respective altitude above ground is shown in red. (b) Local ground truth DEM at the 60$th$ frame. (c) Estimated DEM, corresponding to (b), obtained using \cite{Schoppmann2021multires}. Notice the grainy effect. (d) Estimated DEM using our approach.}
	\label{fig:dem_rmse}
\end{figure}

Fig. \ref{fig:dem_rmse} compares, on the synthetic dataset, the mapping errors of \cite{Schoppmann2021multires} using a pyramid of residuals with Kalman fusion vs.\  our approach using an OMG pyramid. Fig. \ref{fig:dem_rmse}.c shows systematic errors for \cite{Schoppmann2021multires} due to its pyramid update scheme. In this case, using \cite{Schoppmann2021multires}, measurements are added only to the top two layers for depth values greater than 16 m.
Once the altitude drops, measurements (i.e. residuals) start to go to the last layer, which will suppress all measurements stored in the layers above since the final pyramid is obtained by adding up the \textit{cell's pyramid}.
The overall DEM RMSE difference in Fig. \ref{fig:dem_rmse}.a can also be attributed to its sub-optimal fusion of residuals since stored residuals are not updated if the top layer changes in the next frames.

\subsection{Landing detection results}

\begin{figure}[t]
	\centering
	\begin{tabular}{@{}c@{\hspace{3pt}}c@{\hspace{3pt}}c@{}}
	\includegraphics[trim={15cm 25cm 15cm 5.5cm},clip,scale=0.39]{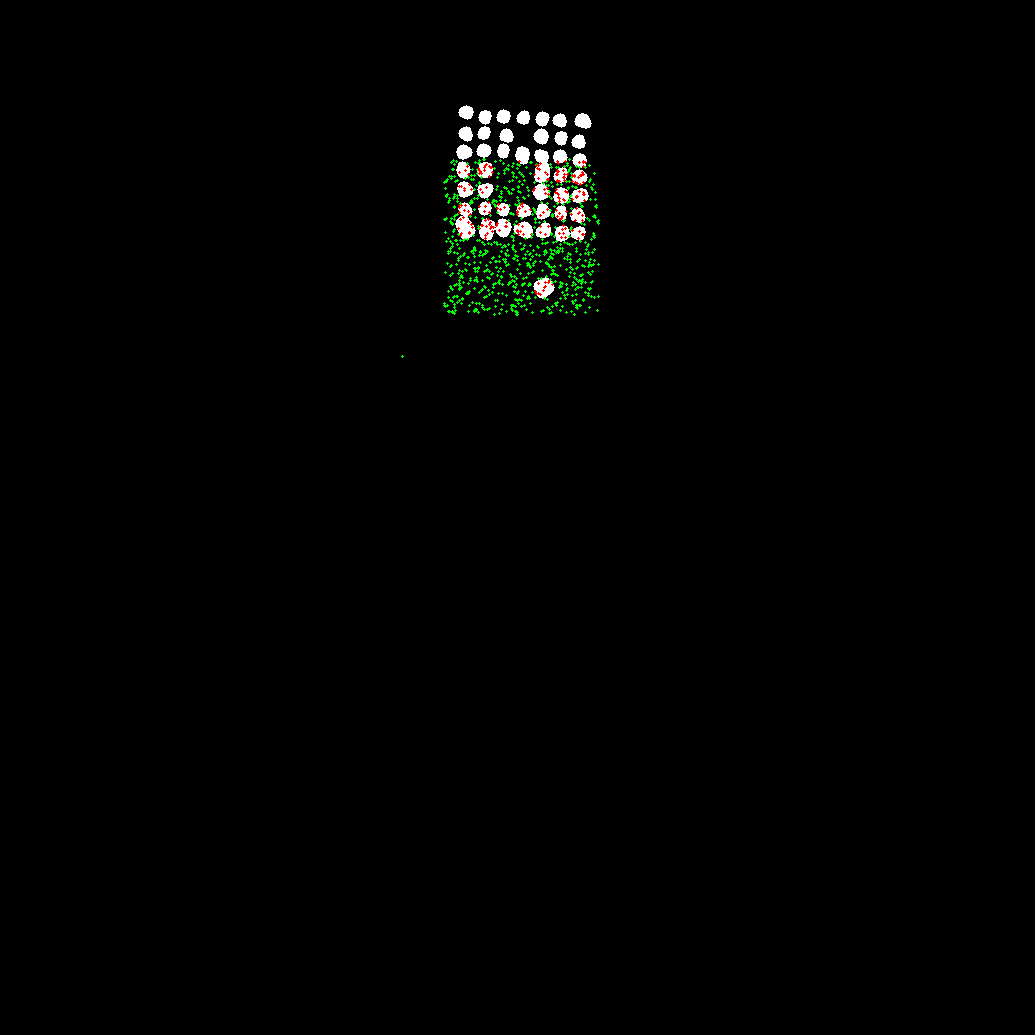} &
	\includegraphics[trim={15cm 25cm 15cm 5.5cm},clip,scale=0.39]{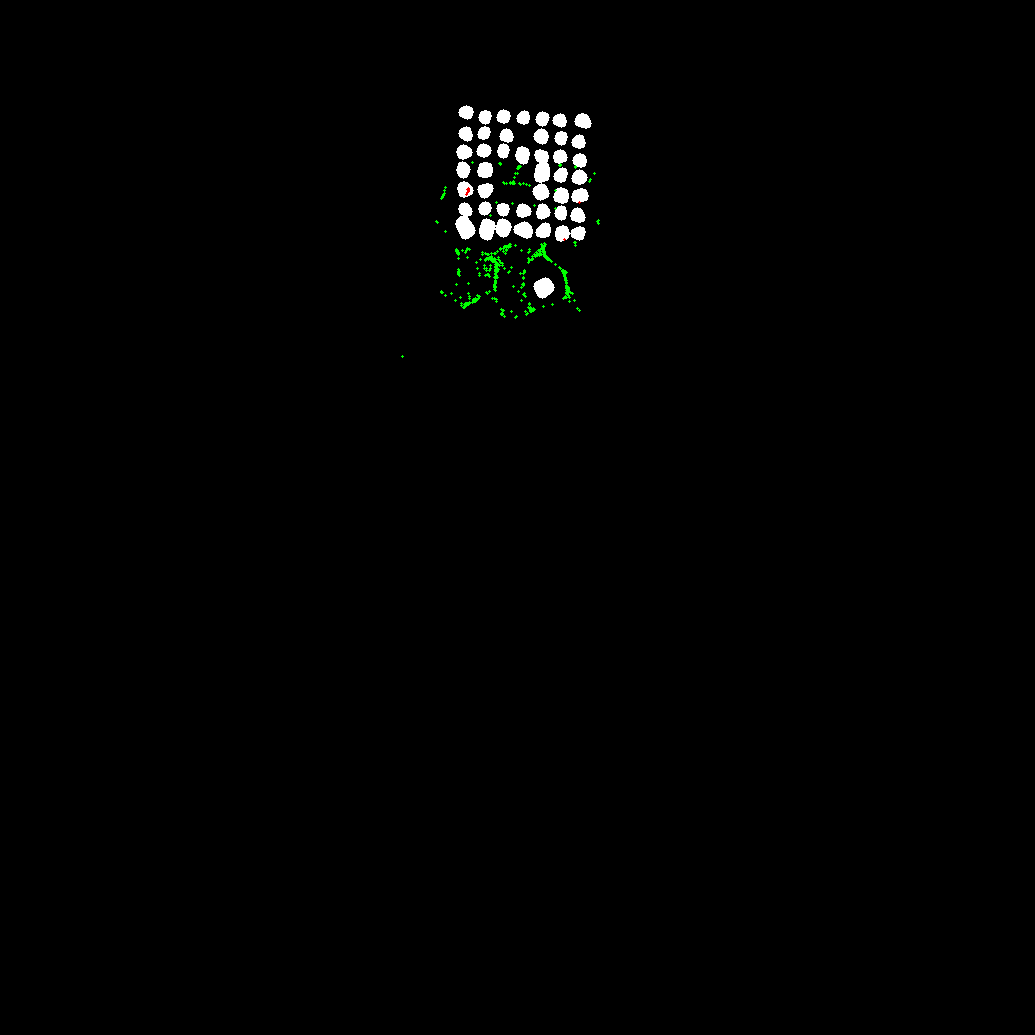} &
	\includegraphics[trim={2.8cm 0.4cm 0.35cm 0cm},clip,scale=1.1]{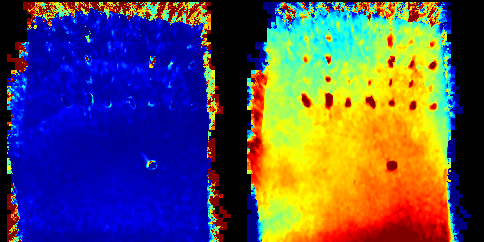} \\
	(a) & (b) & (c)
	\vspace{-7pt}
	\end{tabular}
	\caption{Mean shift applied to 1000 random landing spot samples over a map reconstructed from a Mars Yard flight sequence. Results are overlapped with the annotated rock masks in white.
	(a) Initial samples. (b) Samples after the Mean shift. (c) Estimated DEM used for mean shift.}
	\label{fig:crashes_monte_carlo}
\end{figure}

To validate the mean shift used in our landing spot detection we first performed an experiment, shown in Fig. \ref{fig:crashes_monte_carlo}, where we sampled arbitrary $N=1000$ map locations from reconstructed DEMs of the Mars Yard rock field. and then ran the mean shift for each location. 
We can see that most points on rocks were shifted towards a safer area with exception of one rock where the mean shift found a local minimum. This emphasizes the need for multiple peak detection. \par

Table \ref{tab:crashes_per_resolution} shows the results of continuous landing selection with our full pipeline on two Mars Yard flight sequences shown in Fig. \ref{fig:my_flights}. The DEM size was set to 12$\times$12 m (about the size of the rock grid) to stress the algorithm. As we lower the map resolution, false segmented rocks lead ultimately to selected landing locations on the smaller rocks when using the simple distance transform maximum \cite{Schoppmann2021multires} whereas our proposed landing site detection shows that it can minimize these potential hazardous landing spots. There are situations however where there are not enough safe landing cells (green) to select a good landing spot. We can reject these based on the distance transform.

\begin{table}[tb]
	
	\centering
	\scriptsize{
		\begin{tabular}{llll}
			\hline
			Flight id & Map resolution & Distance transform & Multiple shifted   \\
			& & max \cite{Schoppmann2021multires} & peaks (ours) \\
			\hline
			\#1 & 0.05 m &  0 &  0      \\
			\#1 & 0.1  m  &   32  &    4  \\
			\#1 & 0.2  m  &     65   &     0     \\
			\hline
			\#2 & 0.05 m &    0  &  0      \\
			\#2 & 0.1  m  &   0   &     0  \\
			\#2 & 0.2  m  &   34   &      0     \\
	\end{tabular}}
	\caption{Number of failed selected landing locations (i.e. on rocks) during two flight sequences over the Mars Yard rock field.
	The flying altitudes are respectively 5 and 10 m for flight 1 and 2.}
	\label{tab:crashes_per_resolution}
\end{table}

\subsection{Runtime results}

The runtimes reported here were averaged over one of the Mars Yard flight sequences at 10 m AGL. These were processed using a single core of an Intel Xeon E-2286M on a laptop but in Table \ref{tab:runtimes_per_machine} we also report the runtimes on an ARM architecture: Snapdragon 820, where the whole pipeline works at about 5 fps.
Our bottleneck is still the segmentation even though we were able to speed up this module by a factor of 6.75 using the new rolling buffer, as shown on Table \ref{tab:runtimes_seg}. Fortunately, the landing segmentation and detection do not need to work necessarily at camera frame rate. Thus the system can be split in two threads: one slow thread performing segmentation and detection and a real-time thread performing map updates since these need to keep up with the stereo measurements. The pyramid pooling can also be performed by the backend thread, although its computational cost does not seem significant (see Table \ref{tab:runtimes_mapping}). 

Table \ref{tab:runtimes_mapping} compares the runtimes between using our pyramid pooling approach with single layer updates vs a direct multi-resolution update (described in Section \ref{sec:pooling}). There's a significant difference across different map resolutions (analogous to altitude change). This is more noticeable at 1.25~cm/px: while the direct approach needs to update a total of 21 cells per measurement (i.e. the full \textit{cell's pyramid}), the indirect approach only needs one cell update since all measurements are contained at the top layer. This is then copied down using our cache-friendly down-pooling.

\begin{table}
	\vspace{10pt}
	\centering
	\scriptsize{
		\begin{tabular}{lll}
			\hline
			Module & Intel Xeon E-2286M & Snapdragon 820 (ARM)\\
			\hline
			Map Update & 7 ms  & 35 ms\\
			Pyramid Pooling & 1 ms & 9 ms \\
			Segmentation & 20 ms & 147 ms\\ 
			Detection & <1ms & 4 ms\\
	\end{tabular}}
	\caption{Average runtimes on two processors using our default parameters.}
	\label{tab:runtimes_per_machine}
\end{table}

\begin{table}
	\vspace{10pt}
	\centering
	\scriptsize{
		\begin{tabular}{ll}
			\hline
			Baseline \cite{Schoppmann2021multires} & Optimized Segmentation  \\
			\hline
			 135 ms &  20 ms  \\
	\end{tabular}}
	\caption{Average runtimes of the segmentation module. Optimized Segmentation uses the rolling buffer.}
	\label{tab:runtimes_seg}
\end{table}

\begin{table}
	\vspace{10pt}
	\centering
	\scriptsize{
		\begin{tabular}{lll}
			\hline
			Map resolution & Direct update & Indirect update  \\
			 & (Multi layer) & (Single layer + Pyr. pooling) \\
			\hline
			1.25 cm/px &  31 ms &  18 ms     \\
			2.5 cm/px &  17 ms  &  11 ms  \\
			5 cm/px &  13 ms  &   8 ms    \\

	\end{tabular}}
	\caption{Average mapping runtimes using the pyramid schemes, described in Section \ref{sec:pooling}, for 3 layers during a 10 m altitude fly. We tested 3 different map resolutions, i.e., cell footprint at the lowest layer. Due to our layer selection criteria, using a cell footprint of 1.25 cm makes the measurements update the top layer whereas a cell footprint of 5 cm leads to the bottom layer updates.}
	\label{tab:runtimes_mapping}
\end{table}



\section{Conclusion and Future Work}
This work showed how we can perform efficiently multiresolution height mapping, landing segmentation and detection while
not sacrificing accuracy contrary to past work. There are still open system design questions that can lead to future work depending on mission specifications, e.g., a landing site selection that takes into account path planning and a multi-view stereo for general motion. Moreover, this work used map sizes close to the camera FOV. To expand these it will be necessary to mask unchanged regions to avoid unnecessary segmentation and pyramid pooling. \par
We believe that our proposed OMG framework is relevant beyond this problem. However, we noticed on the rock field dataset that for high altitudes the uncertainty on flat sandy terrain becomes greater than on the small rocks due to the lack of texture. Thus, uncertainty should complement roughness not replace it.
Future work will test on-board landing detection using this framework on a Mars Helicopter surrogate \cite{delaune2020extended}.

\section{Acknowledgments}

The research was carried out at the Jet Propulsion Laboratory, California Institute of Technology, under a contract with the National Aeronautics and Space Administration (80NM0018D0004). 

\section{Appendix}
\label{sec:appendix}
In this section, we demonstrate that the mean of a Kalman update is identical to OMG fusion -- despite different uncertainty.
Let $\{x_1, x_2\}$ be two measurements with their respective uncertainties $\{\sigma_{x_1},\sigma_{x_2}\}$.
Using (\ref{eq:OMG_ori}) we obtain:
\begin{equation}
\mu = \frac{\frac{x_1}{\sigma_{x_1}^{2}}+\frac{x_2}{\sigma_{x_2}^{2}}}{\sigma_{x_1}^{-2}+\sigma_{x_2}^{-2}} 
\end{equation}
Multiplying the top and bottom by $\sigma_{x_1}^2\sigma_{x_2}^2$ gives us the multiplication of Gaussians of a Kalman update:
\begin{equation}
\mu = \frac{x_1\sigma_{x_2}^{2}+x_2\sigma_{x_1}^{2}}{\sigma_{x_1}^{2}+\sigma_{x_2}^{2}} 
\end{equation}

\bibliographystyle{ieeetr} 
{\footnotesize
\bibliography{root}
}

\end{document}